**Learning Polynomial Networks for Classification of Clinical Electroencephalograms**

Vitaly Schetinin[1] and Joachim Schult[2]

[1]Department of Computer Science, University of Exeter, Exeter, EX4 4QF, UK

E-mail: V.Schetinin@ex.ac.uk

[2]TheorieLabor, University of Jena, Jena, D 07740, Germany

E-mail: Joachim_Schult@web.de

*Abstract*—We describe a polynomial network technique developed for learning to classify clinical electroencephalograms (EEGs) presented by noisy features. Using an evolutionary strategy implemented within Group Method of Data Handling, we learn classification models which are comprehensively described by sets of short-term polynomials. The polynomial models were learnt to classify the EEGs recorded from Alzheimer and healthy patients and recognize the EEG artifacts. Comparing the performances of our technique and some machine learning methods we conclude that our technique can learn well-suited polynomial models which experts can find easy-to-understand.

Keywords – Classification, electroencephalogram, polynomial network, Group Method of Data Handling,

## 1. Introduction

Electroencephalograms (EEGs) representing the weak potentials invoked by the brain activity give EEG-experts objective information for analysis and classification (e.g., [1] – [6]). Although EEGs are noise and nonstationary signals varying from one patient to the other in a large range of amplitudes and frequencies, the EEGs have been used to assist clinicians to diagnose such diseases as apnea, Alzheimer, dementia and schizophrenia (e.g., [5] – [7]). To make reliable decisions clinicians have to properly separate neural activity of patients from EEG artifacts caused by electrode noise, eye movement, cardiac, and muscle activities. To do so, they use such methods as independent component analysis (e.g., [1], [8]), regression methods (e.g., [9], [10]), principle component analysis (e.g., [11]), and analysis of outliers (e.g., [12]).

To interpret clinical EEGs, Riddington et *al*. [2] have developed the fuzzy logic system in which they calculated the spectral features of EEG and identified the type of EEG corruptions. These features were extracted by using techniques such as parametric modeling and cross-correlation. Using three feed-forward neural networks, they detect subsequently blinking, eye movement and muscle activities of patients. Finally, the classification was enhanced by incorpo-rating heuristic criteria taking in account a spatial distribution of electrodes on the scalp. These heuristics were implemented as a set of rules in an expert system.

The methods [3] – [6] suggested for classification of EEGs are based on the fully connected feed-forward neural network (FNN) for which users have to properly predefine a suitable network structure as well as a learning method for fitting synaptic weights of the network.

For classification of EEGs recorded from sleeping newborns assigned by medical experts to different risk groups, Breidbach *et al.* [5] have applied a FNN including 72 input neurons and two output neurons with a sigmoid activation function. The learning algorithm that they used maximizes a Euclidean distance between the output vectors belonging to different risk groups. Analyzing the clusters in a space of two output variables, they discovered a correlation between the output vectors and the risk groups and finally found a FNN with 24 input nodes, which provide the better classification despite that the sub-clusters belonging to different risk groups strongly overlap.

The neural network system [6] was developed to assess the dementia of Alzheimer type. It consists of two FNNs: one decides between dementia and non-dementia patients and other estimates severity of the illness. The EEGs recorded via 15 standard electrodes were presented by power spectrums calculated into nine frequency bands. This neural-network system has distinguished between dementia and non-dementia patients and estimated their severity with the acceptable accuracy.

Although such FNNs can learn to classify EEGs well, corruptions of EEGs still lead to ambiguous results because the underlining brain and muscle activities share their characteristics such as wave shape and frequencies [5], [6]. In the meantime, the classification models learnt by FNNs cannot be comprehensible for medical experts due to a large number of synaptic connections [13] – [15]. On the other hand, the rule extraction and decision tree techniques which provide comprehensible rules are based on the trade-off between the complexity and the classification accuracy of the rules (e.g., [13] – [18]).

To achieve high classification accuracy within a FNN framework, users have to preset a well-defined structure of FNNs, namely, the number of input nodes, hidden and output neurons, and assume a proper set of relevant features. In practice this procedure is implemented by a trial-and-error manner and the standard neural network techniques become very expensive computationally.

To overcome shortcomings of the FNNs, Ivakhnenko [19] – [20] has suggested a Group Method of Data Handling (GMDH) capable for inducing neural networks from noise data Gaussian distributed. Based on an evolutionary strategy, the GMDH generates populations or layers of neurons and then trains and selects those neurons which provide the best classification. During learning the GMDH grows the new population of neurons and the number of layers or complexity of the network increases while a predefined criterion is met. Such models can be comprehensively described by a set of short-term polynomials which users may find more observable than the fully connected neural-network classifiers.

Within a GMDH framework, the performance of neurons is evaluated with a specific criterion calculated on the validation subset of data. As a result, the GMDH-type networks have a nearly optimal complexity and they are able to generalize well if the distribution of noise in data can be assumed as Gaussian. In addition, the GMDH-type networks can be comprehensively described by a concise set of short-term polynomials, which are easy-to-observe for experts [20], [21].

Combining of the GMDH and decision tree techniques reveals promising results in detecting EEG artifacts [22]. Within this technique, the GMDH is used for learning the polynomial network (PN) employed to clean training data from the conflicting examples and noisy features useless for the classification. This PN has been induced without assumptions on the distribution of noise presented in data. As a result, the decision trees induced from the cleaned data provide a better generalization ability. However, within a framework of that paper, the effectiveness of learning the PN when the distribution of noise is unknown has not been analyzed.

Based on the fruitful ideas of GMDH in this paper we describe a new technique developed for learning PNs to classify clinical EEGs corrupted by noise. Throughout the paper we give empirical evidence showing that the proposed technique allows inducing the PNs from clinical EEG data without unrealistic assumptions on the distribution of noise presented in data.

In section 2 we describe the EEG data used in our experiments. Then in section 3.1 we present the GMDH algorithm used for learning polynomial neural networks from data whose noise is assumed to be Gaussian, and in section 3.2 we present the learning algorithm capable of learning the PNs from noise data when the distribution of noise is unknown. In section 4 we experimen-tally compare the classification accuracy of the polynomial and neural network techniques on two EEG datasets: the first data were recorder from a Alzheimer patient and a healthy subject and the second were recorded from sleeping newborns. Finally in section 5 we discuss our results.

## 2. The EEG Data

In our experiments we used two types of clinical EEGs transformed to the frequency domain. The first EEGs are the benchmark data recorded from an Alzheimer and a healthy patient via the standard 19-channels C1, …, C19 during 8 seconds [24]. Muscle artifacts were deleted from these data by an expert. Following [5] we calculated the spectral powers into four standard frequency bands: delta (0-4 Hz), theta (4-8 Hz), alpha (8-14 Hz) and beta (14-20 Hz). As the spectral powers were calculated into ½ sec segments with ¼ sec overlapping, each EEG record consisted of 31 segments presented by 76 spectral features. The first 15 segments were used for training and the remaining 16 for testing, so the training and testing data consisted of 30 and 32 EEG segments, respectively.

The second EEGs were recorded from sleeping newborns via the standard EEG channels, C3 and C4, sampled with 100 Hz. The spectral features were calculated at the 10-second segments into six frequency bands such as subdelta (0-1.5 Hz), delta (1.5-3.5 Hz), theta (3.5-7.5 Hz), alpha (7.5-13.5 Hz), beta 1 (13.5-19.5 Hz), and beta 2 (19.5-25 Hz). Additionally for each band the values of relative and absolute powers were calculated. Such values were calculated for channels C3 and C4 as well as for their sum, C3+C4, so the total number of the features was 36. Values of these features were normalized to be with zero mean and unit variance.

Using the additional channels, the EEG-experts have recognized cardiac, eye movement, muscle and noise artifacts and then labeled all segments in the sleep EEGs. In this experiment we used the EEGs recorded from 12 newborns: one EEG was used for training and the remaining 11 for testing. The training EEG contains 1347 labeled segments and one of the testing EEGs contains 808 segments in which the artifact rates are 6.53% and 8.79%, respectively. The remaining 10 testing EEGs we used to evaluate the inter-individual variability of the classification model obtained on the training EEG. These EEGs contain 9406 labeled segments in which the artifact rate is 30.5%, much higher than in the training data.

### 3. GMDH-Type Polynomial Networks

In this section first we describe GMDH algorithm capable of inducing polynomial neural networks from noisy data under the assumption that the noise is Guassian. Second, we present our algorithm for learning PNs from data when the distribution of noise is unknown.

### 3.1. GMDH Algorithms

GMDH-type polynomial networks are the multilayered feed-forward networks consisting of the so-called supporting neurons which have at least two inputs $v_1$ and $v_2$ [19] – [21], [23]. A transfer function $g$ of the neurons is described by short-term polynomials, for example, by a nonlinear polynomial:

$$y = g(\mathbf{v}; \mathbf{w}) = w_0 + w_1 + w_1 v_1 + w_2 v_2 + w_3 v_1 v_2, \tag{1}$$

where $\mathbf{v} = (v_1, v_2)$ is an input vector and $\mathbf{w} = (w_0, w_1, \ldots, w_3)$ is a weight vector of the neuron consisting of coefficients $w_0, w_1, \ldots, w_3$.

Using the supporting neurons, GMDH algorithm builds new generation, or layer, of the candidate-neurons and then selects the best of them. The candidate-neurons are selected with the so-called exterior criterion which evaluates the generalization ability of neurons on the unseen data defined as the validation data. So user has to predefine the exterior criterion as well as the number of neurons, $F$, selected in each layer.

Giving the large $F$, the user increases the chance to find out a global minimum of cross-validation error however the large values of $F$ increase the computational expenses. In practice GMDH algorithms perform enough well for $F$ equal to the number of input variables, $m$. The best performance is achieved for $F = 0.4 * \binom{m}{2}$, where $\binom{m}{2}$ is the number of combinations by 2 from $m$ given inputs [19], [20].

The layers of GMDH-type networks grow up one-by-one. In the first layer the candidate-neurons are connected to inputs $x_1, \ldots, x_m$, and in the next layer they are connected to the outputs of the $F$ neurons selected in the previous layer. Some GMDH algorithms allow also combining between the input nodes and neurons taken from the previous layers [23].

As a given activation function (1) has two arguments $v_1$ and $v_2$, the first layer, $r = 1$, consists of candidate-neurons $y_1^{(1)}, \ldots, y_{L1}^{(1)}$, where $L_1 = \binom{m}{2}$. Having training these neurons, GMDH algorithm selects the $F$ best of them and then generates the next layer in which it generates $L_r = \binom{F}{2}$ candidate-neurons. This procedure of generation and selection of the candidate-neurons cycles and the network grows in size while the value of the exterior criterion decreases.

The exterior criterion can be defined on the training and validation datasets $\mathbf{D}_A$ and $\mathbf{D}_B$ consisting of $n_A$ and $n_B$ examples respectively, $n_A + n_B = n$, $\mathbf{D}_A \cup \mathbf{D}_B = \mathbf{D}$, where $\mathbf{D} = (\mathbf{X}, \mathbf{y}^o)$, $\mathbf{X}$ is a $n \times m$ matrix of the input data and $\mathbf{y}^o$ is a $n \times 1$ target vector. The training data $\mathbf{D}_A$ are used to fit the weight vector $\mathbf{w}$ of the supporting neuron so that to minimize the sum squared error, $e = \sum_k (g(\mathbf{v}^{(k)}; \mathbf{w}) - y_k^o)^2, k = 1, ..., n_A$. The validation data $\mathbf{D}_B$ are used to control the complexity or the number of layers of the GMDH-type network during learning.

For fitting weights $\mathbf{w}$ the conventional GMDH algorithms exploit a least square method which provides the effective estimates of weights if the training data are Gaussian distributed [19] – [21], [23]. For real-world data for which a Gaussian distribution is unrealistic such estimations become biased (e.g., [25]). One way used in such cases to avoid this problem is to make the estimations of weights without unrealistic assumptions about the distribution of training data. Below in section 3.2 we describe our learning method based on this approach.

The complexity of the GMDH-type network is controlled by calculating the value of the exterior criterion, $CR_i^{(r)}$, $i = 1, ..., L_r$, for each candidate-neuron on the whole data $\mathbf{D}$ as follows:

$$CR_i^{(r)} = \sum_k (g_i(\mathbf{v}^{(k)}; \mathbf{w}) - y_k^o)^2, k = 1, ..., n. \tag{2}$$

The value of $CR_i^{(r)}$ as we can see is dependent on how well the $i$th neuron classifies the unseen data $\mathbf{D}_B$. Therefore the value of $CR^{(r)}$ is expected large for the neurons with poor generalization ability and small for the neurons which generalize well.

The values of $CR^{(r)}$ are calculated for all the candidate-neurons in the layer $r$ and GMDH then sorts them in an ascending order, $CR_{i1}^{(r)} \leq ... \leq CR_F^{(r)} \leq ... \leq CR_{Lr}^{(r)}$, so that the first $F$ neurons provide the best classification accuracy. The minimal value of the exterior criterion, $CR_m^{(r)}$, equal to $CR_{i1}^{(r)}$ is used to check the following stopping rule:

$$\left| CR_m^{(r)} - CR_m^{(r-1)} \right| < \Delta, \tag{3}$$

where $\Delta > 0$ is a constant given by user.

This rule is based on an observation that value of $CR_m^{(r)}$ decreases rapidly at the first layers of GMDH-type network and relatively slowly near to an optimal number of layers, and further increasing the number of layers causes increasing the value of $CR_m^{(r)}$ because of over-fitting [19] – [21]. Thus the number of layers in the network increases one-by-one until the stopping rule is met at the layer $r^*$. Subsequently we can take a desired GMDH-type network of a nearly optimal complexity from the $(r^* - 1)$th layer.

### 3.2. Fitting the Neuron Weights

As mentioned above unrealistic assumptions on the distribution of real-world data lead to the biased estimations the neuron weights. However we can use a learning algorithm which is not dependent on the distribution of training data. Below we describe our method.

According to the given transfer function (1), the inputs of the supporting neurons are connected to the pairs of the input variables $(x_i, x_j)$, $\forall i \neq j = 1, \ldots, m$ for the first layer and to the outputs of the neurons $(y_i, y_j)$, $\forall i \neq j = 1, \ldots, F$, for the next layers. So for the training and validation of the supporting neurons, we can denote their input data as the $n_A \times 2$ and $n_B \times 2$ matrices $\mathbf{U}_A$ and $\mathbf{U}_B$, respectively. Using these notations we can describe our learning method as follows.

Initially $k$ is set to zero and the algorithm initiates a weight vector $\mathbf{w}^0$ by random values. At the next step $k$ the algorithm calculates a $n_A \times 1$ error vector, $\mathbf{\eta}_A{}^k$, on the data $\mathbf{D}_A$ as follows:

$$\mathbf{\eta}_A^k = g(\mathbf{U}_A; \mathbf{w}^k) - \mathbf{y}_A^o. \tag{4}$$

On the validation data $\mathbf{D}_B$, it calculates the $n_B \times 1$ error vector $\mathbf{\eta}_B{}^k$:

$$\mathbf{\eta}_B^k = g(\mathbf{U}_B; \mathbf{w}^k) - \mathbf{y}_B^o, \tag{5}$$

as well as the corresponding mean squared error, $e_B(k)$, of the neuron:

$$e_B(k) = 1/n_B \sum_i (\eta_{Bi}^k)^2, i = 1, \ldots, n_B. \tag{6}$$

The error $e_B$ has to be minimized during learning for a finite number of steps $k$. Formally we can complete the learning if the following inequality is met at the step $k^*$:

$$e_B(k^* - 1) - e_B(k^*) < \delta, \tag{7}$$

where $\delta > 0$ is a constant which depends on the level of noise in data $\mathbf{X}$ as well as on the ratio $n_B/n$ given by user.

Until this inequality is met, the current weight vector $\mathbf{w}^{k-1}$ is updated accordingly with the following learning rule:

$$\mathbf{w}^k = \mathbf{w}^{k-1} - \chi \|\mathbf{U}_A\|^{-2} \mathbf{U}_A \mathbf{\eta}_A^{k-1}, \tag{8}$$

where $\chi$ is a given learning rate, and $\| \cdot \|$ is a Euclidean norm.

The desired estimation of weights is achieved for a finite number of steps, $k^*$, if the learning rate $\chi$ lies between 1 and 2, for a proof see [25]. A simple explanation of the above learning rule can be given in a space of two weight components $w_1$ and $w_2$. Let us assume that in this space there is a desirable region $\mathbf{w}^*$ for which condition (7) is met for any vector $\mathbf{w} \in \mathbf{w}^*$. Assume also that at step $k$ a vector $\mathbf{w}^{(k)} \notin \mathbf{w}^*$ as depicted in Fig. 1.

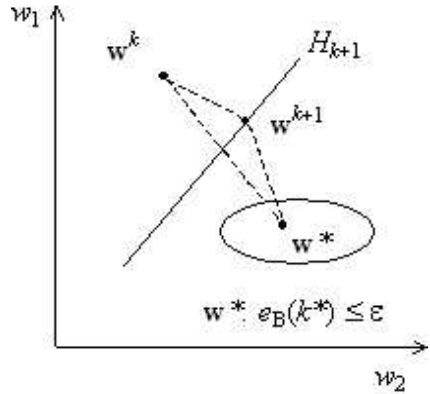

**Fig. 1:** Projection method in a space of two weight components $w_1$ and $w_1$.

Obviously, value $e_B$ is proportional to the distance between the current vector $\mathbf{w}^k$ and the region $\mathbf{w}^*$. Accordingly to rule (6), the new vector $\mathbf{w}^{(k+1)}$ is an orthogonal projection of vector $\mathbf{w}_k$ on hyperplane $H_{k+1}$ located between $\mathbf{w}^k$ and region $\mathbf{w}^*$. We can see that the new vector $\mathbf{w}^{(k+1)}$ is closer to the desirable region $\mathbf{w}^*$ than the previous vector $\mathbf{w}^{(k)}$, and therefore $e_B(k+1) < e_B(k)$. By induction, we can write down that for any $k \leq k^*$, $e_B(k) < e_B(k-1) < \ldots < e_B(0)$ is a monotonically decreasing series.

Varying $\chi$ between 1.0 to 2.0, we can obtain different learning curves for neurons. Fig. 2 shows Residual Squared Error (RSE) of the neuron calculated on the training data set for the values of $\chi$ equal to 1.25, 1.5, 1.75, and 2.0. As we can see, the RSE decreases with a maximal speed when the learning rate $\chi = 2.0$.

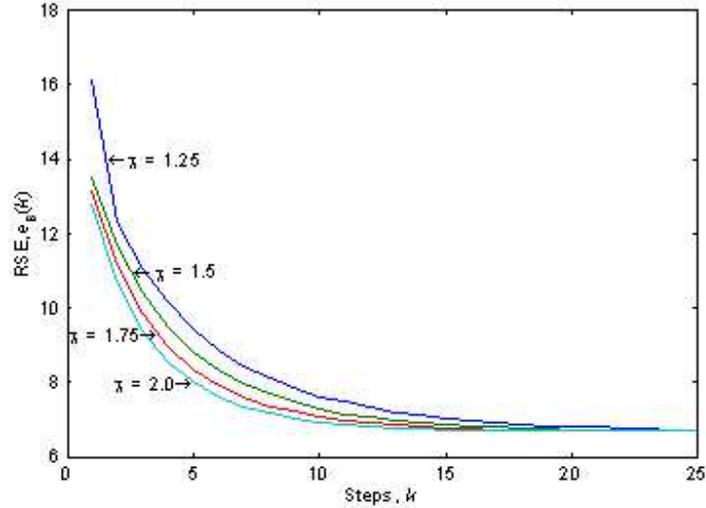

**Fig. 2:** The learning curves were calculated for χ = 1.25, 1.5, 1.75 and 2.0 on the EEG data.

Having given χ = 1.9 in our experiments, we found that the best performance is obtained with a ratio $n_B/n = 0.5$, δ = 0.015 and the initial weights distributed by a Gaussian $N(0, 1)$. In this case the number k* did not exceed 25 steps as depicted in the above Fig. 2.

## 4. Experimental Results

In this section we describe our experiments with the EEGs described in the above section 2. The first experiment is conducted on the benchmark EEG and the second on the EEGs recorded from sleeping newborns.

### 4.1. The Alzheimer Benchmark EEG

In our first experiment the Alzheimer EEG was recorded from one patient only, so that the classification task seems simple because of the absence of the inter-individual variability in the EEG. Moreover, the difference in the EEGs recorded from an Alzheimer patient and a healthy subject can simply reflect the inter-individual variability. Nevertheless, these benchmark EEG data are useful for testing the different classifier's schemes.

Within our PN technique, we defined an activation function (1) and set $F = 1$. As a result, a PN which was induced consists of four input nodes, two hidden nodes presented by variables $y_1^{(1)}$ and $y_2^{(2)}$, and one output node as depicted in Fig. 3.

The induced PN is comprehensively described by a set of three short-term polynomials:

$$y_1^{(1)} = 0.696 + 0.391x_{11} + 0.248x_{69} - 0.231x_{11}x_{69},$$

$$y_1^{(2)} = 0.386 + 0.564y_1^{(1)} + 0.542x_{73} - 0.485y_1^{(1)}x_{73},$$

$$y_1^{(3)} = 0.191 + 0.776\,y_1^{(2)} + 0.238x_{76} - 0.204y_1^{(2)}x_{76},$$

where $x_{11}$ is the delta in channel C11, $x_{69}$, $x_{73}$, and $x_{76}$ are the beta in channels C12, C16, and C19, respectively.

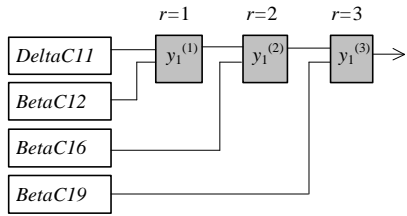

**Fig. 3.** A polynomial network for classifying EEGs of Alzheimer and healthy patients

For comparison we applied a standard neural network technique to these data and found that a FNN using eight principal components and two hidden neurons provides the best classification accuracy. We also applied a conventional GMDH technique to this problem. All three neural networks, the FNN, GMDH, and PN, misclassified one testing segment, i.e., their testing error rate was 3.1%.

### 4.2. The Artifact Detection in Sleep EEGs

In the second experiment we compare the classification accuracy of the standard FNN, GMDH-type and PN techniques on the EEGs recorded from sleeping newborns. Additionally to the performance we evaluated the sensitivity and specificity of the classifiers: the sensitivity is calculated as TP/(TP+FN) and the specificity as TN/(TN+FP), where TP, TN, FP, and FN are the number of patterns classified as true positive, true negative, false positive and false negative, respectively. The performance is calculated as (TP+TN)/(TP+TN+FP+FN). Here positive patterns are associated with artifacts and negative with normal segments.

A specific of the above classification problem is that the PN leans from the data presenting the current EEG segment and the preceding segments are not taking into account. Additional information coming from the other channels has been used only by the experts to label the EEG artifacts.

Fig. 4 depicts the histograms of three EEG features over the training, validation, and testing datasets consisting of 673, 674, and 808 segments, respectively. These features are the absolute powers of subdelta in channels C3 and C4, and absolute power of theta in C4. These histograms show us how different can be the training and testing EEGs recorded from two newborns during sleep hours.

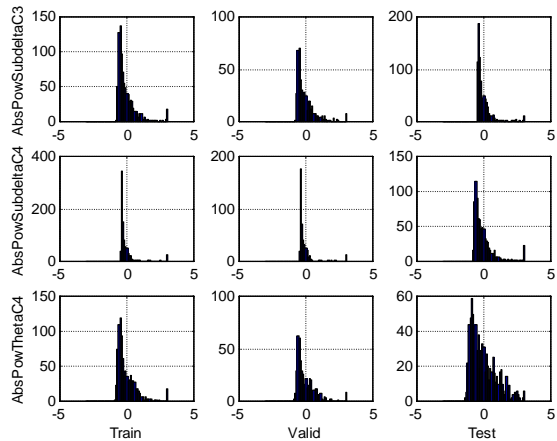

**Fig. 4.** Histograms of three EEG features over the training, validation, and testing datasets

For training the FNNs, we used the standard technique based on the principal component analysis and a fast Levenberg-Marquardt back-propagation learning algorithm provided by MATLAB. The FNNs with a given number of hidden neurons and randomly initialized weights were trained 30 times.

For inducing the GMDH-type network as well as the PN, we used a modified algorithm described in section 3. This algorithm allows the connections between different layers of neurons. It takes a random pair of the neurons and creates a new neuron which is added to the network if $\mu_{new}$ < min($\mu_1$, $\mu_2$), where $\mu_{new}$, $\mu_1$, and $\mu_2$ are the values of criterion (2) calculated for the new and taken neurons, respectively. The growth of the network terminates after a pre-specified number of failed attempts of improving the performance; this number was specified equal to 7. For both networks we used a transfer function (1) and *F* equal 240.

The experimental results averaged over 30 runs are shown in Table 1 which lists the mean and 95% confidence interval of the sensitivity, specificity, and performance calculated for the FNN, GMDH and PN on the single test consisting of 808 EEG segments.

Applying the standard neural network technique, we found out that the FNNs exploiting 10 hidden neurons and 24 principal components provided the best performance Over 30 runs the FNNs correctly classified 95.7±3.2% of the testing data.

**Table 1:** Performance of classifiers on the single test consisting of 808 segments

| # | Classifier | Sensitivity, % | Specificity, % | Performance, % |
|---|-----------|---------------|---------------|---------------|
| 1 | *FNN* | 57.5±37.4 | 99.3±1.4 | 95.7±3.2 |
| 2 | *GMDH* | 63.1±8.0 | 99.9±0.2 | 96.6±0.8 |
| 3 | *PN* | 63.1±13.4 | 99.5±1.0 | 96.3±1.8 |

The GMDH-type networks with the settings described above have correctly classified 96.6±0.8% of the testing data. This is slightly better than that for the FNN.

The PNN has nearly the same performance and correctly classified 96.3±1.8% of the testing data. However for the best PN, the performance was 97.4%. This PN consists of seven neurons connected with the following eight features:

1) *AbsPowSubdeltaC3*, the absolute power of subdelta in channel C3,

2) *AbsPowSubdeltaC4*, the absolute power of subdelta in channel C4,

3) *RelPowThetaC4*, the relative power of theta in C4,

4) *RelPowTheta*, the relative power of theta in C3+C4,

5) *AbsPowThetaC4*, the absolute power of theta in C4,

6) *RelPowAlpthaC4*, the relative power of alpha in C4,

7) *AbsPowAlpha*, the absolute power of alpha in C3+C4,

8) *RelPowBeta2C3*, the relative power of beta2 in C3.

Defining a transfer function (1) as a polynomial function $y = P(v_1, v_2; [w_0 \ w_1 \ w_2 \ w_3])$ in which arguments $v_1$ and $v_2$ are connected either to outputs of the previous neurons or to the features listed above, the best PN can be comprehensively described by a following set of the seven polynomials:

$y_1^{(1)} = P(AbsPowThetaC4, RelPowThetaC4; [0.947 - 0.087 \ 0.073 \ 0.070])$,

$y_2^{(1)} = P(AbsPowSubdeltaC3, RelPowBeta2C3; [0.933 - 0.131 - 0.066 - 0.065])$,

$y_3^{(1)} = P(AbsPowSubdeltaC4, RelPowTheta; [0.932 - 0.204 - 0.008 \ 0.003])$,

$y_4^{(1)} = P(AbsPowAlpha, RelPowAlphaC4; [0.929 - 0.193 \ 0.034 \ 0.036])$,

$y_1^{(2)} = P(y_1^{(1)}, y_2^{(1)}; [0.189 - 0.595 \ 0.666 \ 0.764])$,

$y_2^{(2)} = P(y_3^{(1)}, y_4^{(1)}; [0.250 - 0.003 \ -0.540 \ 1.331])$,

$$y_1^{(3)} = P(y_1^{(2)}, y_2^{(2)}; [0.282 - 0.104 \ 0.045 \ 0.783]).$$

The structure of the best PN consists of eight input nodes, six hidden nodes presented by variables $y_1^{(1)}, \ldots, y_6^{(2)}$ and one output node as depicted in Fig. 5.

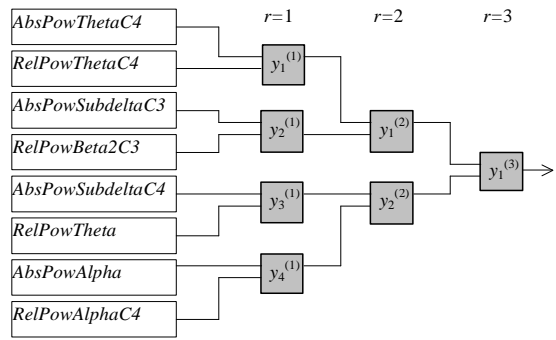

**Fig. 5.** A polynomial network learnt for recognizing EEG artifacts in EEG of sleeping newborns

The inter-individual variability of the above PN can be evaluated in terms of the classification accuracy on the 10 EEGs described in section 2. The classification accuracy on these data was only 68.3%±1.8, that is much less than for a single test. So, the EEG variability over patients is very large and heavily affects the generalization ability of the classifier induced from the EEG recorded from one patient. Clearly, the generalization ability can be improved by learning classifiers from EEGs taken from more than one patients.

Observing the results listed in Table 1 we can conclude that the PN induced by our method performs slightly better than the FNN. The PN reveals the same performance that for the GMDH-type network learnt from the data with a Gaussian assumption on the noise. However the PN learns to classify without any assumption on the distribution of noise in data. In the meantime, the PN is observable for EEG-experts and shows how the features and hidden variables are combined and a decision is arrived.

## 5. Conclusion

Although feed-forward neural networks can learn to classify EEGs well, such classifiers cannot be easily comprehensible for clinicians. On the other hand, the rule extraction and decision tree techniques, which are able to produce the comprehensible rules, are commonly based on the trade-off between the complexity and the classification accuracy of rules (e.g., [13] – [18]). Con-trary to this approach our PN technique aims to induce classification models observable for experts keeping the classification errors down.

Using our technique in our experiments, we have induced two classification models from the clinical EEGs for which the distribution of noise cannot be assumed Gaussian. The first model is for classifying the benchmark EEGs of an Alzheimer and a healthy patient and the second for an automated recognition of cardiac, eye movement, muscle and other artifacts in the EEGs of sleeping newborns. For artifact recognition we used the features calculated for the current EEG segment while additional information coming from other channels outside EEG has been used only by the experts to label the EEG artifacts.

The first practical result obtained in this paper is that the induced polynomial models can be observable for EEG-experts, that is, these models give useful information on that how the decisions are made. The second is that we compared the performance of our polynomial and some machine learning techniques on the EEG data. The third is that the inter-individual variability has been evaluated for the classifier learnt from the EEG recorded from one patient. The final result is that the polynomial model has been learnt for an automated recognition of all the types of EEG artifacts including cardiac, eye movement, muscle, and electrode noise which were visually recognizable for the EEG-experts.

The above results have been obtained on a few EEG records that the EEG-experts could label visually. In practice, the visual labeling of EEGs requires a lot of expertise and human efforts, so that the lack of labeled EEGs is a value problem. Certainly, our results should be confirmed on a large volume of EEGs which EEG-experts will be able to label.

Thus we can conclude that our PN technique is able to provide the observable classification models and at the same time keep their error down without making unrealistic assumptions on the distribution of noise presented in data. We believe that this technique seems promising for mining clinical EEGs.

## Acknowledgments

This work was supported by the University of Jena, Germany, and particularly by the University of Exeter, UK, under EPSRC Grant GR/R24357/01. The authors are grateful to Frank Pasemann for enlightening discussions, Joachim Frenzel and Burkhart Scheidt from the Clinic of the Uni-versity of Jena for the EEG records, to Richard Everson and Jonathan Fieldsend from the University of Exeter for useful comments.

## References

[1] Makeig S, Bell A, Jung T, Sejnowski T (1996) Independent component analysis of electroencephalographic data. In: Touretzky D, Mozer M, Hasselmo M (eds) Advances in neural information processing systems. Cambridge, MA The MIT Press, 145–151

[2] Riddington E, Ifeachor E, Allen E, Hudson N, Mapps D (1994) A fuzzy expert system for EEG Interpretation. In: E. Ifeachor E, Rosén K (eds) Neural Networks and Expert Systems in Medicine and Healthcare (NNESMED'94). Plymouth, England 291-302

[3] Anderson C, Devulapalli S, Stolz E. (1995) EEG signal classification with different signal representations. In: Girosi F, Makhoul J, Manolakos E, Wilson E (eds) Neural Networks for Signal Processing. IEEE Service Center: Piscataway 475-483

[4] Galicki M, Witte H, Dörschel J, Doering A, Eiselt M, Grießbach G (1998) Common optimization of adaptive preprocessing units and a neural network during the learning period. Application in EEG Pattern Recognition. Neural Networks 10: 1153-1163

[5] Breidbach O, Holthausen K, Scheidt B, Frenzel J (1998) Analysis of EEG data room in sudden infant death risk patients. Theory in Biosciences 117: 377-392

[6] Hibino S, Hanai T, Nagata E, Matsubara M, Fukagawa K, Shirataki T, Honda H, Kobayashi T (2000) An Assessment System of Dementia of Alzheimer Type Using Artificial Neural Networks. In: Malmgren H, Borga M, Niklasson L (eds) Artificial Neural Network in Medicine and Biology (ANNIMAB'01). Goteborg, Sweden Springer Verlag

[7] Silvestri-Hobson R (2000) Abnormal Neonatal EEG. E-medicine Inc. Available: http://www.emedicine.com/neuro/topic545.htm

[8] Jung TP, Makeig S, Humphries C, Lee TW, McKeown MJ, Iragui V, Sejnowski TJ (2000) Removing electroencephalographic artifacts by blind source separation. Psychophysiology 37: 163–178

[9] Whitton J, Lue F, Moldofsky H (1978) A spectral method for removing eye-movement artifacts from the EEG. Electroencephalography and Clinical Neurophysiology 44: 735–741

[10] Woestenburg J, Verbaten M, Slangen J (1983) The removal of the eye-movement artifact from the EEG by regression analysis in the frequency domain. Biological Psychology 16: 127–147

[11] Berg P, Scherg M (1991) Dipole models of eye activity and its application to the removal of eye artifacts from the EEG and MEG. Clinical Physics and Physiological Measurements 12: 49–54

[12] Brunner D, Vasko R, Detka C, Monahan J, Reynolds C, Kupfer D. (1996) Muscle artifacts in the sleep EEG: Automated detection and effect on all-night EEG power spectra. Journal of Sleep Research 5: 155–164

[13] Quinlan J (1993) *C4.5: Programs for Machine Learning*. Morgan Kaufmann

[14] Holsheimer M, Siebes A (1994) Data mining: The search for knowledge in databases. Report CS-R9406, CWI Amsterdam, Netherlands

[15] Setiono R (2000) Generating concise and accurate classification rules for breast cancer diagnosis. Artificial Intelligence in Medicine 18: 205-219

[16] Garzes A, Broda K., Gabby D (1998) Symbolic knowledge extraction from trained neural networks. Technical Report TR-98-014, Imperial College London

[17] Hayashi Y, Setiono R, Yoshida K (2000) A comparison between neural network rule extraction techniques for the diagnosis of hepatobiliary disorders. Artificial Intelligence in Medicine 20: 205-216

[18] Towell G, Shavlik J (1993) The extraction of refine rules from knowledge based neural networks. Machine Learning 13: pp. 71-101

[19] Madala H, Ivakhnenko AG (1994) Inductive Learning Algorithms for Complex Systems Modeling. CRC Press Inc.: Boca Raton

[20] Müller JA, Lemke F, Ivakhnenko AG (1998) GMDH algorithms for complex systems modeling. Math and Computer Modeling of Dynamical Systems 4: 275-315

[21] Müller JA, Lemke F (2003) Self-Organizing Data Mining Extracting Knowledge from Data. Trafford Publishing, Canada British Columbia

[22] Schetinin V, Schult J (2004) The combined technique for detection of artifacts in clinical electroencephalograms of sleeping newborns. IEEE Transaction on Information Technologies in Biomedicine (vol. 8) 1: 28-35

[23] Nikolaev NL, Iba H (1999) Automated discovery of polynomials by inductive genetic programming. In: Zutkow J, Ranch J (eds) Principles of Data Mining and Knowledge Discovery (PKDD'99). Springer, Berlin 456-461

[24] The data were collected by D. Duke and K. Nayak from SCRI, Florida State University, and available at http://www.scri.fsu.edu/~nayak/chaos/data.html.

[25] Bondarko VA, Yakubovich VA (1992) The method of recursive goal inequalities in adaptive control theory. Int. Journal of Adaptive Control and Signal Processing, May: 141-160